\documentclass{article}

\PassOptionsToPackage{numbers, compress}{natbib}
\usepackage[preprint]{timeseries_workshop}
\usepackage{multirow}
\usepackage{mathrsfs}

\usepackage[utf8]{inputenc} 
\usepackage[T1]{fontenc}    
\usepackage{hyperref}       
\usepackage{url}            
\usepackage{booktabs}       
\usepackage{amsfonts}       
\usepackage{nicefrac}       
\usepackage{microtype}      
\usepackage{xcolor}         

\title{Beyond LoRA: Exploring Efficient Fine-Tuning Techniques for Time Series Foundational Models}

\author{%
  Divij Gupta\thanks{Corresponding Author. Alt. - divij.can@gmail.com}, Anubhav Bhatti, Surajsinh Parmar \\
  AI Engineering Team, SpassMed Inc. \\ Toronto, Ontario, Canada\\
  \texttt{\{divij.gupta, anubhav.bhatti, suraj.parmar\}@spassmed.ca} \\
}

\begin{document}

\maketitle

\begin{abstract}
Time Series Foundation Models (TSFMs) have recently garnered attention for their ability to model complex, large-scale time series data across domains such as retail, finance, and transportation. However, their application to sensitive, domain-specific fields like healthcare remains challenging, primarily due to the difficulty of fine-tuning these models for specialized, out-of-domain tasks with scarce publicly available datasets. In this work, we explore the use of Parameter-Efficient Fine-Tuning (PEFT) techniques to address these limitations, focusing on healthcare applications, particularly ICU vitals forecasting for sepsis patients. We introduce and evaluate two selective (BitFit and LayerNorm Tuning) and two additive (VeRA and FourierFT) PEFT techniques on multiple configurations of the Chronos TSFM for forecasting vital signs of sepsis patients. Our comparative analysis demonstrates that some of these PEFT methods outperform LoRA in terms of parameter efficiency and domain adaptation, establishing state-of-the-art (SOTA) results in ICU vital forecasting tasks. Interestingly, FourierFT applied to the Chronos (Tiny) variant surpasses the SOTA model while fine-tuning only 2,400 parameters compared to the 700K parameters of the benchmark.
\end{abstract}

\section{Introduction}
Foundational Models (FMs) have advanced deep learning by offering robust, pre-trained architectures effective across multiple domains. Inspired by their success in language and vision, researchers have begun exploring Time Series FMs (TSFMs) to address the complexities of large-scale time series data, such as non-standard quantization, varying scales and frequencies, and irregular intervals \cite{survey,chronos,lagllama}. While TSFMs perform well on datasets similar to their pre-training data (e.g., retail, finance, transport, weather), they underperform in domains with scarce publicly available data like healthcare \cite{divij_lora, midl_peft}. Healthcare data involves diverse signals and tasks—including stress detection \cite{anubhav_stress}, emotion recognition \cite{anubhav_emotion}, vitals forecasting \cite{spass1} and others in both contact and non-contact settings \cite{divij_iot}. Although task-specific models exist, they lack the generalized representations learned by large FMs.Recent efforts to develop domain-specific TSFMs, such as SleepFM \cite{sleepfm} and ECG-FM \cite{ecg-fm}, have resulted in highly specialized models which are, however, less adaptable to a broader range of tasks.

A common method of integrating domain knowledge into FMs is through Parameter-Efficient Fine-Tuning (PEFT), which is widely used in language and vision FMs \cite{peft_survey}. These techniques fine-tune only a small subset of weights—either selected from the existing FM architecture or newly introduced, while keeping the rest of the model weights frozen. This approach leverages the rich feature representations learned by FMs during pre-training while integrating domain-specific knowledge through targeted fine-tuning. By doing so, it reduces the computational burden and inefficiencies associated with fine-tuning the entire FM for a specific task or dataset. However, the impact of PEFT on TSFMs for out-of-domain data remains largely unexplored. A recent study by \cite{divij_lora} provides insights into the application of Low-Rank Adaptation (LoRA) \cite{lora}, a prevalent PEFT technique, within TSFMs in the healthcare domain. There are, however, more recent and efficient PEFT techniques that claim to achieve comparable, if not superior, performance to LoRA while fine-tuning even fewer parameters. 
This is particularly advantageous in scenarios with limited datasets, however they are yet to be applied and benchmarked in the context of TSFMs. To address this gap, we introduce several other PEFT techniques into TSFMs for forecasting vital signs of sepsis patients in critical care and conduct a comparative analysis. Our contributions are summarized as: 

\begin{enumerate} 
    \item We present and assess two selective and two additive PEFT methods, thereby broadening the spectrum of fine-tuning approaches explored with multiple configurations of the Chronos \cite{chronos} TSFM.
    \item We perform a thorough comparative analysis of these PEFT methods against LoRA, highlighting their influence on performance and parameter efficiency in the context of forecasting vital signs, a task involving out-of-domain modalities for TSFMs.
    \item We benchmark our findings against state-of-the-art (SOTA) models that are trained from scratch with a large number of parameters. Our results demonstrate that PEFT can enable certain TSFM variants to achieve, and in some cases exceed, the performance of these models while requiring fine-tuning of significantly fewer parameters. 
\end{enumerate}

\section{Method}

In this section, we provide a brief overview of our methodology, which includes the TSFM used as the backbone for our experiments, the two selective and two additive PEFT methods, the dataset and its preprocessing, and finally, the implementation details.

\noindent \textbf{Backbone:} Recently, multiple TSFMs have been introduced, each employing different strategies for modeling time series data. Among these, we use the Chronos \cite{chronos}  family of TSFMs in our experiments due to their superior performance across various settings—including zero-shot, fully fine-tuned, and LoRA fine-tuned settings \cite{chronos, divij_lora}. Chronos utilizes the encoder-decoder transformer architecture from the T5 \cite{t5} language FM family to perform time series forecasting. The model simplifies data processing by discretizing real-valued inputs through a binning and scaling function. Additionally, Chronos offers multiple-sized model variants providing valuable insights into the effects of scaling when introducing PEFT on larger FMs.

\noindent \textbf{Selective PEFT:} Selective PEFT provides domain adaptation by fine-tuning only a small set of model weights already present in the architecture. For our experiments, we use:

\begin{itemize}
    \item BitFit \cite{bitfit}. This technique focuses on fine-tuning only the bias terms within the model architecture. By adjusting these bias parameters, BitFit can achieve task/data-specific adaptations with minimal changes to the overall model structure.
    \item LayerNorm (LN) Tuning \cite{layer_norm}. This technique fine-tunes the parameters of the LN components present in the attention module of the model architecture. By adjusting the weights and biases of LN layers, the model can learn to better normalize activations across different tasks, potentially stabilizing training and improving performance.
\end{itemize}

\noindent \textbf{Additive PEFT:} Additive PEFT enables domain adaptation by introducing new parameters into the model architecture, which are then fine-tuned keeping all the original weights frozen. For our experiments, we use the recently introduced:

\begin{itemize}
    \item Vector-based Random Matrix Adaptation (VeRA) \cite{vera}. This technique adds a parallel processing path in the linear layers of the attention module using fixed, randomly initialized rank matrices, similar to LoRA. These matrices are shared across layers and remain frozen during training. VeRA introduces two learnable scaling vectors that, when multiplied by the frozen matrices, facilitate domain adaptation. The weight update is given by: 
    \begin{equation}
        W' = W + \lambda_b B \lambda_d A.  
    \end{equation}
    Here, $W$ is the original weight matrix, $W'$ is the updated weight matrix, $B$ and $A$ are the frozen rank matrices, and $\lambda_b$ and $\lambda_d$ are the learnable scaling vectors (used as diagonal matrices) for matrices $B$ and $A$ respectively that modulate the adaptation.
    
    \item Fourier Transform for Fine-Tuning (FourierFT) \cite{fourier}.
    This technique leverages the Fourier transform to reduce the number of trainable parameters by encoding them into a small set of spectral coefficients. These coefficients correspond to a subset of spectral entries, which are randomly initialized and shared across all layers. The inverse discrete Fourier transform (IDFT) is then applied to convert the modified spectral data back into the spatial domain, forming the weight update matrix. The weight update in FourierFT can be expressed as:
    \begin{equation}
        W' = W + \alpha  \mathscr{R} \{ \mathscr{F}^{(-1)}(E \odot c) \}
    \end{equation}
    where $E$ represents the frozen spectral entries, $c$ is the set of learnable spectral coefficients, $\odot$ denotes the element-wise operation to create the spectral matrix, $\mathscr{F}^{(-1)}$ is the IDFT, $\mathscr{R}$ extracts the real part of the transformed data, and $\alpha$ serves as a scaling factor.

\end{itemize}

\begin{table}
\scriptsize

\caption{Results compare MeanBP and HR forecasts across various fine-tuning settings, including full fine-tuning (FullFT) as reported in \cite{divij_lora}. We also analyze the total number of trainable parameters (in millions) to assess efficiency. The best results for each fine-tuning setting within each TSFM are underlined, while the overall best values are highlighted in bold.}

\centering
\begin{tabular}{c|c|ccc|ccc|c}
\hline

\multirow{2}{*}{Model}  & \multirow{2}{*} {Setting} & \multicolumn{3}{c}{MeanBP*}  & \multicolumn{3}{c}{HR*} & \#Params.\\
& & MSE $\downarrow$   & DTW  $\downarrow$  & MAPE  $\downarrow$ & MSE $\downarrow$  & DTW  $\downarrow$  & MAPE $\downarrow$ & (in M.) $\downarrow$    \\

\hline
\hline

\multirow{3}{*}{\begin{tabular}[c]{@{}c@{}}Bhatti et al. \cite{spass1}\end{tabular}} 
    & N-HiTS    & 19.81  & \textbf{16.32} & 7.90  & \underline{7.18}  & \textbf{7.92} & \underline{6.27}  & 0.7\\
   & N-BEATS  & 27.42  & 18.60  & 9.64  & 12.98  & 17.90  & 9.15  & 0.73\\
   & TFT  & \textbf{19.00}  & 23.46  & \underline{7.78} & 7.57  & 15.79 & 6.52 & 0.53 \\  \hline

\multirow{7}{*}{\begin{tabular}[c]{@{}c@{}}Chronos\\ (Tiny)\end{tabular}}  
    & 0-shot    & 25.60  & 21.40  & 8.64  & 7.37  & 11.41  & 5.97  & - \\
   & Full FT \cite{divij_lora}      & 19.90  & 20.51  & 8.03  & 8.80  & 14.99 & 6.86 & 8.3  \\
   & LoRA \cite{divij_lora}   & 19.79 & 19.86 & 7.90  & 7.22  & 11.17 & 5.90 & 0.049  \\ 
   & BitFit       & 20.68 & 19.79 & 8.01 & 7.35 & 11.31  & 5.95 & \textbf{0.0002} \\
   & LN Tuning  & 20.25 & 19.05 & 7.87 & 7.38 & 11.30  & 5.94 & 0.005  \\
   & VeRA & 20.16 & \underline{18.50} & 7.80 & 7.29 & 11.19  & 5.96 & 0.013  \\ 
   & FourierFT & \underline{19.51} & 18.55 & \textbf{7.76} & \textbf{7.06} & \underline{10.80}  &  \textbf{5.81} & 0.002  \\ \hline
   
\multirow{7}{*}{\begin{tabular}[c]{@{}c@{}}Chronos\\ (Small)\end{tabular}} 
    & 0-shot    & 25.04  & 20.28  & 8.49  & 7.19  & 11.01  & \underline{5.92} & - \\
   & Full FT \cite{divij_lora}       & 20.93  & 20.95 & 8.23  & 10.04 & 16.53  & 7.37 & 46.1 \\
   & LoRA \cite{divij_lora}     & 19.89  & 20.44  & 8.02  & \underline{7.08} & 10.83  & 5.88 & 0.147  \\ 
      & BitFit       & 20.65 & 19.87 & 8.09 & 7.19 & 10.90  & 5.93 & \underline{0.0005} \\
   & LN Tuning  & 19.81 & 19.85 & 7.99 & 7.21 & 10.88  & 5.94 & 0.016  \\
   & VeRA & 20.98 & \underline{19.08} & 7.98 & 7.21 & 10.87  & 5.94 & 0.038  \\ 
      & FourierFT & \underline{19.65} & 19.98 & \underline{7.94} & 7.20 & \underline{10.82}  & 5.93 & 0.003  \\ \hline
   
\multirow{7}{*}{\begin{tabular}[c]{@{}c@{}}Chronos\\ (Base)\end{tabular}}  
    & 0-shot    & 25.70 & 20.32  & 8.53  & 7.33  & 11.02  & 5.96 & -  \\
   & Full FT \cite{divij_lora}       & 20.80  & 21.09 & 8.20  & 10.15 & 16.82  & 7.37 & 201 \\
   & LoRA \cite{divij_lora}      & 20.12  & 21.06  & 8.06  &\underline{7.25}  & \underline{10.96}  & \underline{5.93} & 0.442  \\  
      & BitFit       & \underline{19.77} & 19.30 & 7.91 & 7.42 & 11.16  & 5.99 & \underline{0.0007}  \\
   & LN Tuning  & 19.87 & 19.58 & 7.92 & 7.41 & 11.16  & 5.99 & 0.047 \\
   & VeRA & 21.58 & 18.30 & 7.92 & 7.40 & 11.10  & 5.97 & 0.113  \\ 
      & FourierFT & 20.98 & \underline{17.96} & \underline{7.86} & 7.40 & 11.12  & 5.97 & 0.007  \\ \hline

\end{tabular}
\label{full}\\
\footnotesize *MSE values are normalized by 1e-4, and DTW values by 1e-3 for clearer interpretation.
\end{table}

\noindent \textbf{Dataset and Processing:} We used the publicly available eICU Collaborative Research Database \cite{eicu} to forecast mean blood pressure (MeanBP) and heart rate (HR) for sepsis patients in ICU. Following methods from \cite{divij_lora, spass1}, missing values were imputed with forward filling. We extracted 9-hour windows of vital signs before diagnosis, with the first 6 hours as context and the last 3 hours as the prediction horizon. Vitals were sampled every 5 minutes, resulting in context and horizon windows of 72 and 36, respectively. A low-pass filter was then applied to reduce noise, followed by global min-max scaling for normalization. The preprocessed dataset consisted of 4,020 samples from 1,442 patients, split into training, validation, and test sets in an 8:1:1 ratio, with no patient data overlapping across splits.

\noindent \textbf{Implementation Details:} 
We optimized performance across various data types and fine-tuning configurations using learning rates between 1e-2 and 1e-5 with the Adam optimizer. For additive PEFT methods, we fine-tuned all attention weight matrices (Query, Key, Value, and Output) for both FourierFT and VeRA for fair comparison with LoRA used in \cite{divij_lora}. For VeRA, we set the adapter rank $r$ to 16, which is much higher than the rank 2 used for LoRA in prior work \cite{divij_lora}, following recommendations \cite{vera} due to the smaller fraction of fine-tuned weights. For FourierFT, $\alpha$ was set to 300, with number of coefficients, $n$ tuned to 50.  We have provided a rationale for selecting the above values for $r$ and $n$ in Section \ref{appendix}. Other hyperparameters were kept as specified in the original model checkpoints. Experiments were conducted in PyTorch on an NVIDIA RTX 4090 GPU, with forecasting performance evaluated using mean squared error (MSE), dynamic time warping (DTW), and mean average percentage error (MAPE) as per \cite{spass1, divij_lora, spass2}. Owing to the probabilistic nature of the forecasts, the median of 20 samples is used for each prediction, with metrics averaged over 10 runs for robust evaluation.

\section{Results and Discussion}
\label{results}

We evaluate the effectiveness of various PEFT techniques across different Chronos model variants—Tiny, Small, and Base—comprising 8.3M, 46.1M, and 201M trainable parameters, respectively, with the results presented in Table \ref{full}. These findings are further compared against LoRA and full model fine-tuning as reported in \cite{divij_lora}. We observe that, in nearly all configurations, the PEFT methods consistently outperform the TSFM without domain adaptation (zero-shot) and full fine-tuning approaches. Techniques such as BitFit, LN Tuning, VeRA, and FourierFT demonstrate comparable or superior performance to LoRA across multiple metrics for both MeanBP and HR forecasting, while requiring significantly fewer trainable parameters. Notably, FourierFT, when applied to the Chronos (Tiny) variant, surpasses the SOTA model \cite{spass1}. Our calculations, based on the configuration in their study, the SOTA model comprises approximately 700K trainable parameters, whereas FourierFT requires fine-tuning of only 2,400 parameters. This highlights the effectiveness of training a small set of parameters for domain adaptation, leveraging the strong generalization capabilities of large FMs, rather than training smaller models with a narrow focus on specific data or tasks.

Among all the PEFT techniques, BitFit fine-tuned the smallest number of parameters while delivering comparable and sometimes even the best performance. For example, in the case of Chronos (Base), which originally has 201M parameters, BitFit fine-tuned only 768 parameters and achieved the best values for the MSE metric in MeanBP forecasting. It closely followed FourierFT, the best-performing PEFT in that setting, which fine-tuned 7200 parameters. After BitFit, FourierFT consistently gave the best results for both vitals across multiple metrics in various settings. It excelled particularly with the other two TSFMs particularly with Chronos (Tiny) which surpassed the SOTA by fine-tuning a small number of parameters compared to techniques like LoRA, VeRA, and LN Tuning. This aligns with the findings of the original work \cite{fourier}, where the authors propose that only a small number of sparse coefficients need to be learned in the spectral domain to achieve the desired effect in the spatial domain. LN Tuning required more fine-tunable parameters compared to FourierFT, but still produced comparable results in several settings. Among the new PEFT techniques introduced, VeRA fine-tuned the most parameters, approaching the parameter count of LoRA. This can be explained by the way PEFT is performed in VeRA. Since VeRA fine-tunes only the scaling vectors, the rank of the frozen matrices needs to be much higher than in LoRA to achieve similar results. For instance, while LoRA used rank-2 matrices to fine-tune 49K parameters for Chronos (Tiny), VeRA required rank-16 matrices to fine-tune 13K parameters to attain comparable performance. Interestingly, we observe that for Chronos (Base) using LoRA with 442K trainable parameters yields the best performance for forecasting HR vitals. This can be attributed to the relatively low variability in HR readings, which likely necessitate tuning a larger number of parameters in a model of this size (201M original parameters) that has been pre-trained on a broad range of highly variable time series data. 

Overall, the results underscore a critical trade-off between the number of fine-tuned parameters and model performance. While techniques such as BitFit and FourierFT demonstrate that competitive outcomes can be achieved by fine-tuning a minimal subset of parameters, methods like VeRA and LN Tuning involve fine-tuning a larger number of parameters, which can lead to improved performance in certain settings but at the cost of increased computational requirements. Moreover, additive methods like FourierFT, VeRA, and LoRA offer greater flexibility, owing to the additional hyperparameters that accompany these techniques.

\section{Conclusion and Future Work}
In this work, we conducted a comparative study of various PEFT techniques within the context of adapting TSFMs for forecasting healthcare time series data, specifically vital signs in ICUs. We explored two selective PEFT methods, BitFit and LayerNorm Tuning, alongside two additive PEFT methods, VeRA and FourierFT. Our results showed that in addition to the commonly used LoRA method, other PEFT techniques achieve comparable or superior performance, fine-tuning only a small fraction of the total model parameters. Notably, FourierFT, when applied to Chronos (Tiny) TSFM, outperformed SOTA models trained from scratch on this task. Our experiments further demonstrated that tuning as few as 256 parameters (BitFit) can yield competitive results underscoring the importance of continued research in the evolving TSFM space. In future work, we plan to investigate hybrid approaches that combine multiple PEFT techniques and explore more complex TSFMs as the field progresses.


\bibliographystyle{myver}
\small
\bibliography{refs}

\appendix

\section{Appendix}
\label{appendix}
In the appendix, we present the experiments supplementary to our work. Herein we include the results from tuning the hyperparameters of the additive PEFT techniques, and also the experiment results from the other two Chronos variants (Mini - 20.4M parameters and Large - 708M parameters) for comprehensiveness.


\subsection{VeRA}
We observe in Table \ref{vera} that setting the rank $r$ to 16 provided the optimal results when tuning this hyperparameter. Additionally, as $r$ increased, the change in the number of tunable parameters was not substantial. This is due to VeRA's approach, where instead of learning full matrices, the PEFT learns only scaling vectors, significantly reducing the parameter overhead despite higher ranks.

\begin{table} [!h]
\scriptsize
\caption{Results comparing MeanBP and HR forecasts across different ranks $r$ of the introduced adapters by VeRA for adapting Chronos (Tiny).}

\centering
\begin{tabular}{c|ccc|ccc|c}
\hline

\multirow{2}{*} {Rank (r)} & \multicolumn{3}{c}{MeanBP*}  & \multicolumn{3}{c}{HR*} & \#Params.\\
 & MSE $\downarrow$   & DTW  $\downarrow$  & MAPE  $\downarrow$ & MSE $\downarrow$  & DTW  $\downarrow$  & MAPE $\downarrow$ & (in M.) $\downarrow$    \\

\hline
\hline 
    1  & 20.43 & 21.31 & 8.16 & 7.36 & 11.31 & 5.99 & \textbf{0.0123} \\
   2    & 20.20 & 20.76 & 8.06 & 7.30 & 11.20 & 5.96 &  0.0124  \\
   4    & \textbf{20.10} & 18.65 & 7.87 & 7.31 & \textbf{11.18} & 5.96 & 0.0125\\ 
   8    & 20.25 & 18.59 & 7.87 & 7.32 & 11.20 & 5.96 & 0.0127    \\
   \textbf{16}   & 20.16 & \textbf{18.50} & \textbf{7.80} & \textbf{7.29} & 11.19  & 5.96 & 0.0131 \\
   32  & 20.26 & 18.59 & 7.87 & 7.35 & 11.24 & 5.96 & 0.0139 \\ \hline

\end{tabular}
\label{vera} \\
\footnotesize *MSE values are normalized by 1e-4, and DTW values by 1e-3 for clearer interpretation.
\end{table}

\subsection{FourierFT}

Similar to VeRA, we observe in Table \ref{ff} that setting the number of spectral coefficients $n$ to 50 yielded the optimal results when tuning this hyperparameter. This configuration consistently provided the best performance across the evaluated settings.

\begin{table}[!h]
\scriptsize
\caption{Results comparing MeanBP and HR forecasts across different number of spectral coefficients $n$ of the introduced adapters by FourierFT for adapting Chronos (Tiny).}

\centering
\begin{tabular}{c|ccc|ccc|c}
\hline

\#Spectral & \multicolumn{3}{c}{MeanBP*}  & \multicolumn{3}{c}{HR*} & \#Params.\\
Coefficients ($n$) & MSE $\downarrow$   & DTW  $\downarrow$  & MAPE  $\downarrow$ & MSE $\downarrow$  & DTW  $\downarrow$  & MAPE $\downarrow$ & (in M.) $\downarrow$    \\

\hline
\hline 
    25     & 19.89 & 18.67 & 7.85 & 7.17 & 10.98 & 5.88 & \textbf{0.0012}\\
    \textbf{50}  & \textbf{19.51} & \textbf{18.55} & \textbf{7.76} & \textbf{7.06} & \textbf{10.80}  &  \textbf{5.81} & 0.0024 \\
    
   100  & 19.71 & 20.44 & 7.93 & 7.31 & 11.19 & 5.99 &  0.0048  \\
   200   & 19.85 & 20.46 & 7.99  & 7.39 & 12.01 & 6.16 & 0.0096 \\ \hline

\end{tabular}
\label{ff} \\
\footnotesize *MSE values are normalized by 1e-4, and DTW values by 1e-3 for clearer interpretation.
\end{table}

\subsection{Chronos}

\begin{table}[!h]
\scriptsize
\caption{Results comparing MeanBP and HR forecasts across different fine-tuning settings including full fine-tune (FullFT) from \cite{divij_lora} for Chronos (Mini) and Chronos (Large). The best values for the fine-tune settings are underlined for each TSFM.}
\centering
\begin{tabular}{c|c|ccc|ccc|c}
\hline

\multirow{2}{*}{Model}  & \multirow{2}{*} {Setting} & \multicolumn{3}{c}{MeanBP*}  & \multicolumn{3}{c}{HR*} & \#Params.\\
& & MSE $\downarrow$   & DTW  $\downarrow$  & MAPE  $\downarrow$ & MSE $\downarrow$  & DTW  $\downarrow$  & MAPE $\downarrow$ & (in M.) $\downarrow$    \\

\hline
\hline

\multirow{7}{*}{\begin{tabular}[c]{@{}c@{}}Chronos\\ (Mini)\end{tabular}}  
    & 0-shot    & 25.21  & 20.44  & 8.52  & 7.44  & 11.20  & 5.97 & - \\
   & Full FT \cite{divij_lora} & 20.50 & 21.80 & 8.19 & 10.46 & 17.45  & 7.50 & 20.4       \\
   & LoRA \cite{divij_lora}   & \underline{20.05}  & 20.72 & 8.03  & \underline{7.26}  & \underline{10.88} & \underline{5.90} & 0.086  \\ 
   & BitFit     & 20.68  &  20.02 & 8.01 & 7.47 & 11.25 & 5.98 & \underline{0.0005}   \\
   & LN Tuning & 20.48 & 21.26 & 8.19 & 7.48 & 11.22 & 5.99 & 0.008 \\
   & VeRA  & 20.21 & \underline{17.95} & \underline{7.73} & 7.44 & 11.19 & 5.97 & 0.024  \\ 
   & FourierFT  & 20.43 & 21.46 & 8.17 & 7.45 & 11.10 & 5.97 & 0.002 \\ \hline
   
\multirow{7}{*}{\begin{tabular}[c]{@{}c@{}}Chronos\\ (Large)\end{tabular}} 
   & 0-shot     & 25.54 & 19.75  & 8.50  & 7.21  & 10.84 & 5.93 & - \\
   & Full FT \cite{divij_lora} & 21.01  & 20.58  & 8.22  & 9.56  & 16.45  & 7.32 & 708      \\
   & LoRA \cite{divij_lora}    & \underline{20.00}  & 20.34  & 8.06 & \underline{7.16} & \underline{10.76} & \underline{5.91} &  1.1 \\ 
   & BitFit   & 25.09 & 19.49 & 8.43 & 7.30 & 10.92 &  5.94 & \underline{0.001}     \\
   & LN Tuning & 25.09  & 19.49 & 8.43  & 7.30 & 10.92 &  5.94 & 0.125  \\
   & VeRA & 25.09  & 19.49  & 8.43 & 7.30 & 10.92 &  5.94  & 0.3   \\ 
   & FourierFT & 22.11 & \underline{18.24} & \underline{7.97} & 7.31 & 10.91 & 5.94 & 0.014  \\ \hline
      
\end{tabular}
\label{large} \\
\footnotesize *MSE values are normalized by 1e-4, and DTW values by 1e-3 for clearer interpretation.
\label{chronos}
\end{table}

We present the results of experimenting with the remaining Chronos variants in Table \ref{chronos}, where similar trends to those discussed in Section \ref{results} are observed. For MeanBP, VeRA, and FourierFT outperformed LoRA on DTW and MAPE metrics. However, for the lower variability vital, HR forecasting, LoRA remained the best-performing PEFT method. Additionally, for Chronos (Large), fine-tuning fewer parameters than LoRA did not lead to better performance with results stagnating regardless of the number of parameters fine-tuned, likely due to the large-scale nature of the model. FourierFT, however, stood as an exception, yielding distinct outcomes compared to the other newly introduced PEFT methods. This divergence can be attributed to FourierFT's unique processing and learning mechanisms in the spectral domain.

\end{document}